\pdfoutput=1

\documentclass[11pt]{article}

\usepackage[final]{acl}

\usepackage{times}
\usepackage{latexsym}
\usepackage{amsmath}
\usepackage{graphicx}
\usepackage{booktabs}
\usepackage{multirow}
\usepackage[T1]{fontenc}

\usepackage[utf8]{inputenc}
\usepackage{amsthm, amssymb}
\usepackage{microtype}

\usepackage{inconsolata}
\usepackage{mathletters}
\usepackage{graphicx}
\usepackage[inkscapelatex=false]{svg}

\title{Graph-Assisted Culturally Adaptable Idiomatic Translation for Indic Languages}

\author{Pratik Rakesh Singh, Kritarth Prasad, Mohammadi Zaki \and Pankaj Wasnik \\
        Media Analysis Group, Sony Research India\\
        \texttt{\{pratik.singh, kritarth.prasad, mohammadi.zaki, pankaj.wasnik\}@sony.com}}

\begin{document}
\maketitle
\begin{abstract}
Translating multi-word expressions (MWEs) and idioms requires a deep understanding of the cultural nuances of both the source and target languages. This challenge is further amplified by the one-to-many nature of idiomatic translations, where a single source idiom can have multiple target-language equivalents depending on cultural references and contextual variations. Traditional static knowledge graphs (KGs) and prompt-based approaches struggle to capture these complex relationships, often leading to suboptimal translations. To address this, we propose IdiomCE, an adaptive graph neural network (GNN) based methodology that learns intricate mappings between idiomatic expressions, effectively generalizing to both seen and unseen nodes during training. Our proposed method enhances translation quality even in resource-constrained settings, facilitating improved idiomatic translation in smaller models. 
We evaluate our approach on multiple idiomatic translation datasets using reference-less metrics, demonstrating significant improvements in translating idioms from English to various Indian languages.
\end{abstract}

\begin{figure}
    \centering
    \includegraphics[width=1\linewidth]{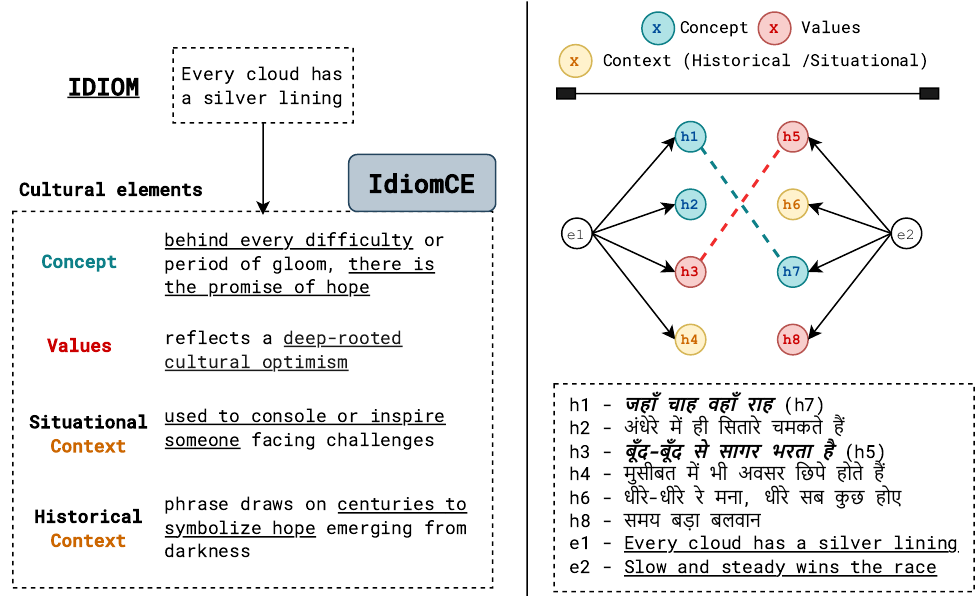}
    \caption{An example of cultural enhanced graph with different cultural elements: Concepts, Values, Context (Historical/Situational) and how we can create relationship among source and target nodes using their cultural elements.}
    \label{fig:toy_example}
\end{figure}

\section{Introduction}
In linguistic terms, \textit{idiom} is a \textit{multi-word expression} (MWE) whose meaning cannot be derived from the literal meanings of its individual parts. Idioms have key properties such as noncompositionality, fixedness, and cultural specificity \cite{Nunberg1994-NUNI}. They are integral to everyday language, enhancing expressiveness and communicative vividness. They often originate from diverse cultural, historical, and situational contexts, making them unique to specific languages or regions \cite{cultural_idm, cultural_idm1}. 

With advancements in large language models (LLMs), neural machine translation (NMT) has significantly improved in handling complex linguistic phenomena, which led to research interest in complex linguistic tasks such as translating idioms across multiple languages \cite{idiomkb, persion-eng-idm, castaldo-monti-2024-prompting}. However, despite these advancements, idiomatic translation remains a major challenge due to the inherent properties of idioms. Traditional NMT systems, both statistical and neural, struggle with noncompositionality, as they primarily process text at the word or phrase level rather than capturing an idiom’s holistic meaning. This often leads to literal translations, distorting the intended meaning of the source text \cite{baziotis-etal-2023-automatic, raunak-etal-2023-gpts, dankers-etal-2022-transformer}.

Recent efforts to address idiomatic translation have primarily relied on (1) idiom dictionary-based substitution \cite{salton-etal-2014-evaluation} and (2) prompting techniques, such as chain-of-thought (CoT) reasoning or explicitly providing figurative meanings in prompts \cite{castaldo-monti-2024-prompting, li2023translatemeaningsjustwords, rezaeimanesh2024comparativestudyllmsnmt}. Although these methods have shown improvements in idiomatic translation, they still fail to overcome key challenges. As shown in Figure \ref{fig:toy_example}, these methods often overlook cultural factors that shape idioms and influence their mappings across languages (\textit{Challenge I}). 
Additionally, they fail to address the one-to-many nature of idioms, where a single source-language idiom may have multiple valid translations in the target language, with the optimal choice depending on the source sentence’s context \cite{persion-eng-idm} (\textit{Challenge II}). 
Moreover, knowledge graph (KG)-based approaches are inherently constrained by the availability of idiom resources, leading to translation gaps when encountering idioms not present in the KG \cite{KG_survey} (\textit{Challenge III}). These challenges pose a critical research question:\\
\textit{How can cultural nuances be effectively integrated into many-to-many idiomatic translation to enhance model performance?}

To address this challenge, one possible approach is to first analyze the cultural dependencies of idioms and identify the specific cultural elements that shape idiomatic expressions across languages. Recent studies in NLP \cite{liu2024_cultureAware} \cite{pawar2024surveyculturalawarenesslanguage} introduce a comprehensive taxonomy of cultural and sociocultural elements, highlighting the need for culturally adaptive models as well as efforts to incorporate cultural awareness. However, even with a structured understanding of these cultural elements, capturing their intricate relationships and effectively leveraging them for one-to-many idiomatic translation remains a significant challenge.

This paper introduces \textbf{IdiomCE}, an inductive graph-based approach that models the relationships between source and target idioms by leveraging complex cultural element mappings, as illustrated in Figure \ref{fig:toy_example}, where source is an English idiom and target are Hindi idioms.
Using link prediction, our method facilitates one-to-many idiomatic translation while preserving cultural relevance across languages. Furthermore, IdiomCE is adaptable, enabling the translation of unseen idioms by leveraging the inductive capabilities of GNNs, effectively addressing the limitations of noisy and limited knowledge bases. Our key contributions are summarized as follows: 
\begin{itemize}
    \item We propose a \textit{cultural element-based data creation} method that generates multiple target idioms for a given source idiom.
    \item We develop an Inductive GNN trained on this graphical data, leveraging link prediction to enable one-to-many idiomatic translation (addressing \textit{Challenge I} and \textit{II}).
    \item We design an adaptable pipeline that extends to unseen idioms using the inductive capabilities of GNNs (addressing \textit{Challenge III}).
    \item Using English as a pivot language, we extend our approach to facilitate idiomatic translation across Indic languages without needing to train GNN models between all possible pairs of languages.
\end{itemize}

\section{Related work and Motivation}
\subsection{Related Works}
\textbf{Idiomatic Text Translation.} Previous studies have explored various strategies to enhance NMT performance for idiomatic translation. \cite{salton-etal-2014-evaluation} introduced a substitution-based method, where source-side idioms are replaced with their literal meanings before translation and reinstated post-translation to improve accuracy. \cite{zaninello-birch-2020-multiword} demonstrated that augmenting training data with idiomatic translations enhances model performance on both source and target sides. Beyond direct translation techniques, researchers have focused on learning non-compositional embeddings and automatically identifying idioms, as explored by \cite{weller2014distinguishing}, \cite{hashimoto-tsuruoka-2016-adaptive}, and \cite{tedeschi-etal-2022-id10m}. More recently, prompting techniques and Chain-of-Thought (CoT) reasoning have been investigated in Large Language Models (LLMs) for idiomatic translation \cite{castaldo-monti-2024-prompting, rezaeimanesh2024comparativestudyllmsnmt}. IdiomKB \cite{idiomkb} further introduced a contextual approach, using figurative meanings as additional context to improve translation quality in LLMs. 


\noindent\textbf{Idiomatic Translation for Indic languages.} Indic languages exhibit significant linguistic diversity and deeply rooted cultural nuances, making idiomatic translation a complex challenge. Despite this, research on idiomatic translation in the Indic language setting remains limited. \cite{shaikh-2020-determination} proposes Idiom Identification using grammatical rule based approach.\cite{9154112} proposes a identification of Gujarati idioms and translation of them using contextual information. \cite{agrawal-etal-2018-beating} present a multilingual parallel idiom dataset encompassing seven Indian languages and English. While these studies offer valuable contributions, the challenge of many-to-many idiomatic translation across Indic languages remains largely under-explored.

\subsection{Motivation}\label{sec:motivation}
\noindent\textbf{Motivation for Cultural significance in Idioms.}
As discussed previously, most of the past studies either use a dictionary-based approach for idiom translation, which is a one-to-one mapping, or provide \textit{figurative meaning} of the idiomatic expression for meaningful translation. Although these approaches appear to perform well, they fail to account for the cultural dependency of idioms, which is deeply embedded within them. This raises the question of how idioms can be effectively mapped from one language to another while considering this cultural dependency. Cultural dependency can be linked to various features, as discussed in \cite{liu2024_cultureAware} and \cite{pawar2024surveyculturalawarenesslanguage}. Identifying these features that influence translation between languages can contribute to the development of more culturally appropriate idiomatic mappings from a source language to a target language.

\noindent\textbf{Motivation for Using GNNs.}
Using a static Knowledge Graph (KG) or dictionary-based approach poses several challenges, which a Graph Neural Network (GNN)-based architecture can effectively address:

\noindent\textit{Limited Generalization.} KGs store only predefined idiomatic translations as edges between nodes, making them incapable of inferring translations for new idioms unless explicitly added. In contrast, GNNs learn graph patterns, enabling them to predict idiomatic translations even for unseen idioms.

\noindent\textit{Lack of Semantic Connectivity.} KGs treat nodes independently, failing to capture relationships between idioms with similar meanings unless explicitly modeled. GNNs leverage neighborhood structures and embeddings, allowing them to infer new translations by recognizing semantic similarities.

\noindent\textit{Polysemy Handling.} KGs require multiple nodes to represent idioms with multiple meanings, increasing complexity. GNNs disambiguate meanings using context, leveraging neighborhood information and learned representations to differentiate between senses based on connectivity.

\section{Methodology}
In this section, we first present the problem statement followed by the training and inference of our methodology, which we call \textbf{IdiomCE}. 

\subsection{Problem Formulation:}

\noindent We address the challenge of replacing idioms in a source language with culturally aware and contextually appropriate multi-word expressions in the target language. Let $\cS$ and $\cT$  denote the sets of graph nodes representing source and target idioms, respectively. The combined set $\cS \cup \cT$ defines the node set $\cV$ in our framework, where each node $v\in \cV $ corresponds to an idiom.

\noindent Each Idiom $v$ is embedded with cultural elements, reflecting its historical, situational, or value-based significance, indicating its relevance to a specific language. Our goal is to identify the most relevant set of target-language idioms $\{\bar{v}: \bar{v}\in \cT\}$ that correspond to a given source-language idiom $v$. We denote this relationship with an edge $e_{v,\bar{v}}$. Let the set of all such edges be $\cE\equiv\{e_{v,\bar{v}}: v\in \cS, \bar{v}\in \cT\}$

\noindent Once we construct or estimate the graph $\cG\equiv(\cV, \cE)$, we use it to generate translations that are both contextually and culturally relevant. Given a sentence in the source language, our approach leverages this graph $\cG$ to produce a culturally and semantically appropriate idiomatic translation in the target language.

\begin{figure*}
    \centering
    \includegraphics[width=0.9\linewidth]{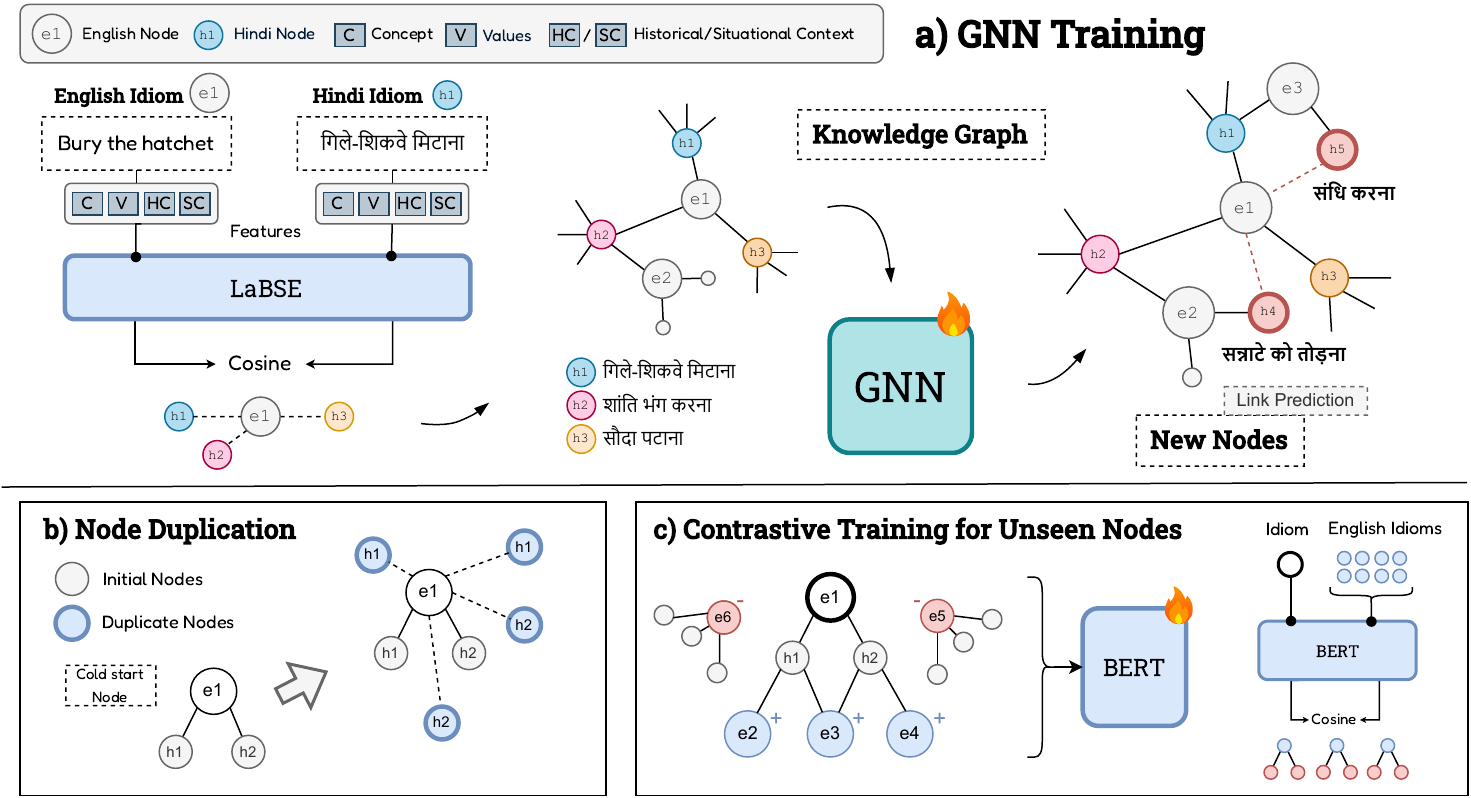}
    \caption{\textbf{Overall training process of IdiomCE}: (a) GNN training – illustrating the creation of a Knowledge Graph using source and target idioms, specifically for \texttt{en-hi}, leveraging LaBSE embeddings and training a GNN for the Link Prediction (LP) task; (b) Node Duplication – demonstrating how we address the cold start issue by duplicating target nodes; and (c) Contrastive Training – showing the training through positive and negative samples and the process of mapping unseen nodes to relevant target idioms.}
    \label{fig:Methodology_figure}
\end{figure*}


\subsection{Training}
In this section, we outline the process of constructing the initial dataset for training our IdiomCE encoder and decoder, followed by the training methodology. An overview of the entire training process is illustrated in Figure \ref{fig:Methodology_figure}.

\noindent {\bfseries GNN Dataset Formation.} We begin by extracting idioms from the collected dataset, as detailed in Section 4 (Datasets), and obtain monolingual idiom sets for each language. For each idiom, we extract three key cultural elements: \textit{Concepts}, \textit{Values}, and \textit{Situational and Historical Context}. These elements are generated using the LLaMA-3.1-405B model and defined based on the Taxonomy of Culture outlined in \cite{liu2024_cultureAware}. Our observations suggest that these elements are highly distinguishable and effectively capture key cultural and sociocultural dimensions essential for mapping English idioms to their counterparts in other languages. The prompt used for generating these cultural elements is provided in Appendix \ref{sec:Generation_prompt}.


\noindent To construct the \textbf{Knowledge Graph (KG)}, we first convert the generated cultural elements into Embeddings (we call it cultural features) with Language-agnostic BERT Sentence Embedding (LaBSE) model \cite{LaBSE}. Once the cultural features for each idiom are generated, we compute the cosine similarity between the cultural features of English and target (Indic) language idioms to establish pairwise mappings, as illustrated in Figure \ref{fig:Methodology_figure}. Moreover, to identify the most relevant idiom pairs for the KG, we focus on outliers within the cosine similarity scores, as these indicate strong semantic relationships. Outlier detection is performed by calibrating thresholds based on the skewness and kurtosis of the data, leveraging both the Inter-Quartile Range (IQR) and $z$-score. By carefully selecting thresholds in these approaches, we ensure that only high-similarity idiom pairs are connected, effectively capturing the most significant relationships. This approach, grounded in robust statistical techniques \cite{ChandolaAnomaly}, ensures that the graph reflects the most salient semantic connections.\\
\noindent As a result of this process, multiple KGs are constructed, each linking English idioms to idioms in a specific Indic language. Formally, each KG is represented as $\cG\equiv (\cV,\cE)$, where $\cV$ denotes the feature of each idiom/node and $\cE$ represents the edges connecting source and target idioms.

\subsection{IdiomCE}
The proposed IdiomCE follows the widely used encoder-decoder architecture for GNN-based link prediction \cite{kipf2016variationalgraphautoencoders} \cite{schlichtkrull2017modelingrelationaldatagraph} \cite{zhao2022learningcounterfactuallinkslink} where a GNN encoder learns node representations, and a decoder predicts link existence probabilities for each node pair. Below, we provide a detailed discussion of the training process for our method.

\noindent{\bfseries Node Duplication Augmentation.}
Once the above KG is constructed, we could encounter the cold start problem due to the sparsity of the dataset, which consists of only a few thousand idioms. This issue arises when certain nodes have few or no connections, leading to under-representation in the GNN during the downstream tasks \cite{hao2020pretraininggraphneuralnetworks, zhang2023coldwarmnet}. To mitigate this, we employ a Node Duplication strategy \cite{guo2024nodeduplicationimprovescoldstart}, which enhances node connectivity and improves representation learning.

\noindent We provide a detailed explanation of our node duplication procedure. Let $\cS$ and $\cT$ represent the sets of source and target language idioms, respectively. For any node $v\in \cV \equiv \cS\cup \cT$, we define its set of neighbors as:
\[\cN_v:=\{\bar{v}: e_{v,\bar{v}} \text{ or }e_{\bar{v},v}\in \cN_v\},\]
where $\cN_v$ consists of all nodes $\bar{v}$ connected to $v$ by an edge.
We extend the methodology of \cite{guo2024nodeduplicationimprovescoldstart} by categorizing source nodes into two types:

\noindent\textit{Cold nodes} ($\cT_{cold}$): Target nodes with fewer than $\delta$ neighbors.\\
\noindent\textit{Warm nodes} ($\cT_{warm}$): Target nodes with at least $\delta$ neighbors. For our experiment we consider $\delta$ equals 3

\noindent For each cold node $v$, we duplicate its neighbors $\cN_v$ and create new corresponding source nodes. We then insert edges from $v$ to these duplicated source nodes, as illustrated in Figure \ref{fig:Methodology_figure}. In this way, we obtain an augmented graph $\cG'$ with these newly created nodes and edges added to the original graph. This approach differs from \cite{guo2024nodeduplicationimprovescoldstart}, where the authors duplicate source nodes directly based on their degree. In contrast, we duplicate source nodes based on the degree of their corresponding target nodes. This strategy enhances the sampling of under-represented cold nodes by leveraging their connections to source nodes.


\noindent {\bfseries IdiomCE Encoder.} As discussed in the previous section, once our augmented $\cG'$ is created, we convert $\cG'$ into the GNN training format by creating a feature vector of each idiom node with a BERT-based embedding model, i.e., LaBSE \cite{LaBSE}. We then construct an initial bi-directional adjacency matrix of edge indices required for training. To ensure generalization across potentially unseen idioms, we employ an inductive GNN for training, specifically SAGEConv \cite{hamilton2018inductiverepresentationlearninglarge}. In SAGEConv, each node updates its representation by aggregating the features of its neighbors. The aggregation is done using a permutation invariant function. In our case, we use the mean aggregator, which computes the average of the feature vectors of a node’s neighbors. This ensures that the order of neighbors does not affect the result. For a given node $v$, let $\mathcal{N}(v)$ represent the set of neighbors and $h_{u}$ denote the features vectors of node $u$. The mean aggregator is defined as:

\begin{equation}
\mathbf{h}_{\mathcal{N}(v)} = \frac{1}{|\mathcal{N}(v)|} \sum_{u \in \mathcal{N}(v)} \mathbf{h}_u.
\end{equation}

\noindent Next, the node's updated representation is computed by concatenating its own feature vector \( \mathbf{h}_v \) with the aggregated neighbor features and then applying a learnable linear transformation followed by a non-linear activation function as given below:

\begin{equation}
\mathbf{h}_v' = \sigma \left( \mathbf{W} \cdot \text{CONCAT}\big(\mathbf{h}_v, \mathbf{h}_{\mathcal{N}(v)}\big) \right).
\end{equation}



\noindent {\bfseries IdiomCE Decoder.}
We perform the task of link prediction by pairing our IdiomCE encoder with a Multi-Layer Perceptron (MLP) model as a decoder. Given a source node $i$ with GNN embeddings $h_{i}$ and target node $j$ with GNN embeddings $h_{j}$ from the Encoder, we first concatenate their embeddings, then pass it through the MLP layer.
\begin{align*}
z_{ij} &= \left[ h_i \, \Vert \, h_j \right],\\
\hat{y}_{ij} &= \operatorname{MLP}(z_{ij}).
\end{align*}

\noindent Once we obtain the prediction from the MLP layer, we then backpropagate using BCE loss and jointly train the GNN and MLP layer for the Link prediction task defined by the loss function given below:

\begin{equation}
\small
\mathcal{L} = -\frac{1}{N} \sum_{(i,j) \in \mathcal{D}} \left[ y_{ij} \log \hat{y}_{ij} + (1-y_{ij}) \log \left(1-\hat{y}_{ij}\right) \right].
\end{equation}


\subsection{Dealing with Unseen nodes}

One of the key properties of inductive GNNs is their ability to generalize to unseen nodes, such as idioms absent from the training set. To incorporate an unseen idiom into a trained GNN, it must be connected to relevant neighbors, allowing the model to compute meaningful node embeddings through message passing. A naïve approach is to add edges by randomly selecting target nodes from the initial dataset. However, this often results in dispersed and suboptimal embeddings due to the lack of semantic coherence in the connections. Therefore, to generate high-quality embeddings for an unseen idiom, it is essential to establish connections with semantically relevant neighbors that closely align with its ideal (gold) translation. Given the one-to-many nature of idioms where a single target idiom may correspond to multiple source idioms conveying the same figurative meaning, it is crucial to connect the unseen node to the most similar target idiom neighbors.\\
To achieve this, we propose training a BERT-based encoder (denoted as $\cB_{CL}(\cdot)$) in a contrastive learning setting \cite{cohan-etal-2020-specter, ostendorff-etal-2022-neighborhood}. The training process leverages a triplet framework designed to align with the graphical structure of our GNN, i.e., $\langle$\textit{anchor } $a$, \textit{positive }$p$, \textit{negative }$n$ $\rangle$ where $a$ denotes the source node representing the idiom in the source language, $p$ denotes the source language nodes that are connected to the anchor (i.e., first-hop neighbors in our KG), and $n$ represents nodes that are disconnected (no path exists) to the anchor, ensuring that they do not share semantic similarity. This triplet construction is used in a contrastive loss $\cL_t$ that minimizes the distance between the anchor and its positive examples while maximizing the distance to the negative examples. Formally, if $h_a$, $h_p$, and $h_n$ are representations of anchor, positive and negative examples, respectively, then with margin $\alpha$,

\begin{small}
\[\mathcal{L_{t}} = \sum_{(a, p, n) \in \mathcal{D}} \max(0, \norm{h_a - h_p} - \norm{h_a- h_n} + \alpha).\]
\end{small}

\subsection{Inference}

\begin{figure}
    \centering
    \includegraphics[width=1\linewidth]{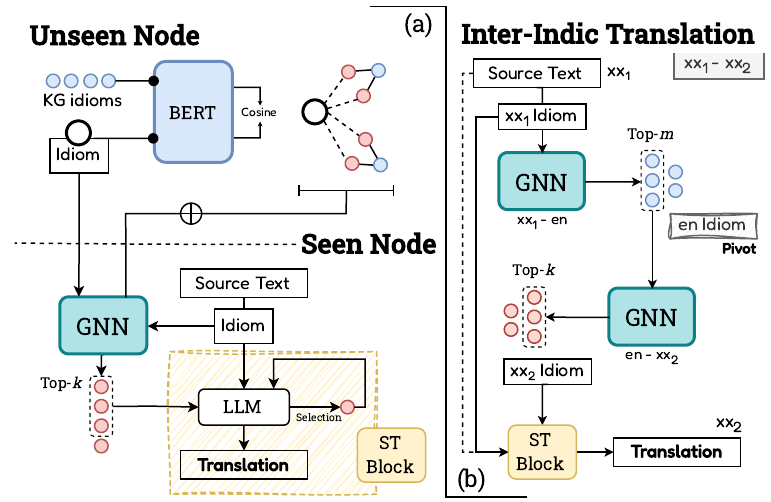}
    \caption{Inference strategy: (a) Unseen \& Seen Node Translation – a BERT-trained GNN adapts to unseen nodes, with the Selection and Translation (ST) block selecting idioms via an LLM before translation; (b) Inter-Indic Translation – using English as a pivot between $xx_1$ and $xx_2$.}
    \label{fig:seenunseen}
\end{figure}

From the trained bi-directional GNNs on English and specific Indic languages, we explore idiomatic translation through three approaches, as illustrated in Figure \ref{fig:seenunseen}: seen nodes, unseen nodes and inter-indic. The \textit{seen nodes}, refer to idioms for which GNN has prior knowledge, including their relationships with other idioms. On the other hand, \textit{unseen nodes} pertain to idioms for which the GNN has no prior information nor any established connections to other idioms. Lastly, \textit{inter-indic} translation where english idioms are treated as pivot, more explanation in section \ref{sec:inter-indic}. We assume idiom detection is a well-explored problem, enabling us to focus directly on the translation task without treating idiom identification as an intermediate step. We also presume that the idiom in the source sentence is provided for retrieval through IdiomCE.
\subsubsection{Seen Nodes}
To infer with \textit{seen nodes}, we first retrieve top-\textit{k} target idioms using the trained GNN by link prediction by providing source idiom as input. Next, we refine the selection by filtering out the most contextually relevant target idiom based on the source sentence. This is achieved by passing the retrieved idioms into a selection prompt as context in an LLM. Finally, once the most relevant target idiom is identified, we perform LLM-based inference by passing the source text, source idiom, and the selected target idiom into a translation prompt. The details of both prompts are provided in Appendix \ref{sec:Selection_prompt} and \ref{sec:Translation_prompt}.

\subsubsection{Unseen Nodes}

For unseen nodes, completely isolated idioms would yield no meaningful results. To address this, we make the following assumption about the training dataset $\cD$.

\noindent\textbf{Assumption.} For any unseen node $u$, $\exists v \in \cD  $ such that $\cos(\cB_{CL}(u),\cB_{CL}(v)) \geq \tau$, where $\tau\in[0,1]$. For our experiments, we choose $\tau$ to be 0.75.
 
To infer on unseen nodes, we first retrieve the most similar idioms in the source language using cosine similarity based on embeddings from the trained contrastive embedding model $\cB_{CL}$. After selecting the top \textit{M} source language idioms, we randomly select five target-language idioms linked to these source idioms and connect them to the unseen idiom, incorporating them into our graphical data. Once integrated, we perform link prediction on the unseen node to retrieve the most suitable target idiom.

\subsubsection{Inter Indic Languages translation} 
\label{sec:inter-indic}
We train the IdiomCE encoder bidirectionally between English and individual Indic languages. In addition to direct translation from $\cS$ to $\cT$, we propose leveraging trained GNNs for indirect translation. Let $\cA_1$, $\cA_2$ and $\cA_3$ be nodes in languages $A_1$, $A_2$ and $A_3$ respectively. Let $\cG_{12}:\cA_1\to \cA_2$ and $\cG_{23}:\cA_2\to\cA_3$ be GNNs trained between the respective languages. To generate a translation from $A_1$ to $A_3$, we use $A_2$ as the \textit{pivot} language, shown in Figure \ref{fig:seenunseen}. 

\section{Experimental set up}

\begin{figure}
    \centering
    \includegraphics[width=1\linewidth]{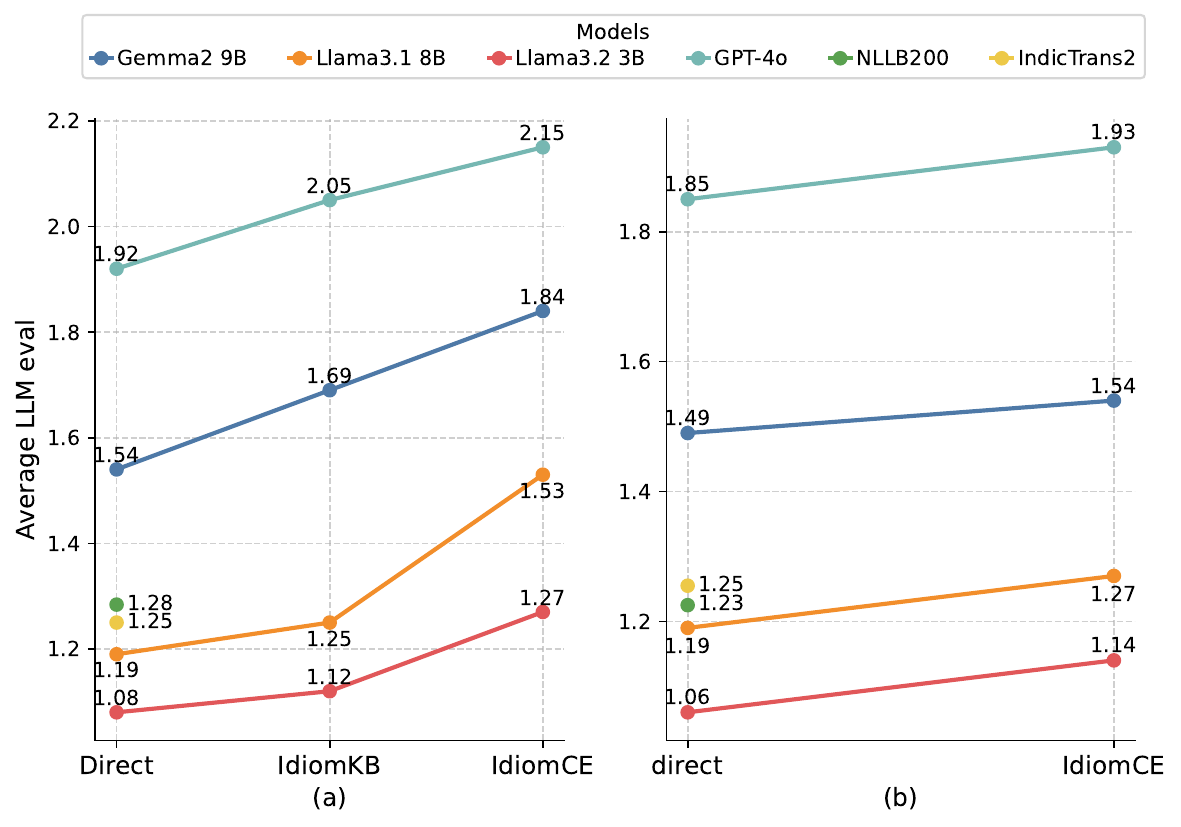}
    \caption{Performance comparison on average LLM score of Models on seen nodes (idiom) (a) and unseen nodes (b) across en-xx direction.}
    \label{fig:unseen-seen-graph}
\end{figure}

{\bfseries Datasets.} 
The initial knowledge graph (KG) construction is based on the dataset from Agrawal et al. (2018) \cite{agrawal-etal-2018-beating}, which provides mappings of idioms between English (en) and seven Indian languages. For our study, we utilize four Indic languages: Tamil (ta), Telugu (te), Bengali (bn), and Hindi (hi). Additionally, we incorporate a parallel idiomatic sentence dataset from Thakre et al. (2018) \cite{thakre2018idiomatic}. Beyond these existing resources, we also web-scraped to collect idioms in various Indic languages. For evaluation, we sample 400 sentences from the MAGPIE dataset \cite{haagsma-etal-2020-magpie} to assess translation effectiveness from English to Indic languages. To analyze performance under different conditions, we conduct experiments in two setups: (1) Seen Idioms, where idioms present in the training data are tested, and (2) Unseen Idioms, where idioms not encountered during training are evaluated. For the Inter-Indic language setting, we curate a dataset of 200 idiomatic sentences per Indic language from the Samanantar dataset \cite{ramesh2023samanantarlargestpubliclyavailable}, ensuring coverage across multiple language pairs, more details on dataset can be found in Appendix \ref{sec:data_comp}.

\noindent {\bfseries Evaluation Metrics.} 
Most automatic evaluation metrics, like BLEU \cite{papineni-etal-2002-bleu,sacrebleu} and ChrF \cite{popovic-2015-chrf}, rely on reference matching but struggle with one-to-many translation, especially idioms, where $n$-gram matches fall short. They also fail to distinguish literal from figurative translations. While CometKiwi \cite{rei-etal-2022-cometkiwi} improves on traditional metrics by being reference-less and semantic-focused, it still struggles to reward high-quality idiomatic translations. Hence, for our evaluation, we adopt the GPT-4o-based evaluation method proposed by \cite{idiomkb} as our primary metric, as it is an LLM-based approach specifically designed for assessing idiomatic translations we call it here LLM-eval and use WMT22-CometKiwi-DA as a supplementary evaluation metric.

\noindent {\bfseries Models.} We test the effectiveness of our approach by using base LLMs of varying sizes like Gemma2 9B \cite{gemmateam2024gemma2improvingopen}, Llama-3.1 8B, Llama-3.2 3B \cite{grattafiori2024llama3herdmodels} and GPT-4o mini \cite{openai2024gpt4technicalreport} in our methodology. We also evaluate our method by comparing them with translations generated from traditional NMT systems like NLLB 3.3B \cite{nllbteam2022languageleftbehindscaling} and IndicTrans2 \cite{gala2023indictrans2highqualityaccessiblemachine}. In our experiments \texttt{Direct} represents either directly prompting the LLM to translate the given source sentence, or passing the sentence through the NMT model for generating translation prompt can we referred from Appendix \ref{sec:direcet_prompt}. Specific training details and performance of GNN and MLP layer with other experimental parameters are added in Table \ref{tab:gnn_performance} in Appendix.

\begin{table*}[ht]
    \centering
    \resizebox{\textwidth}{!}{
        \begin{tabular}{c  |c  |cc  |cc  |cc  |cc}
            \toprule
            \multirow{2}{*}{\textbf{Model}} & \multirow{2}{*}{\textbf{Methods}} & 
            \multicolumn{2}{c|}{{\texttt{en-hi}}} & 
            \multicolumn{2}{c|}{{\texttt{en-bn}}} & 
            \multicolumn{2}{c|}{{\texttt{en-ta}}} & 
            \multicolumn{2}{c}{{\texttt{en-te}}} \\
            \cmidrule(lr){3-4} \cmidrule(lr){5-6} \cmidrule(lr){7-8} \cmidrule(lr){9-10}
            & & \textbf{LLM-eval}& \textbf{COMET} & \textbf{LLM-eval}& \textbf{COMET} & \textbf{LLM-eval}& \textbf{COMET} & \textbf{LLM-eval}& \textbf{COMET} \\
            \midrule
            \multirow{1}{*}{\textbf{NLLB-200}} 
                & \texttt{Direct}& 1.3& 0.70& 1.43& 0.769& 1.18& 0.691& 1.1& 0.643\\
            \midrule
            \multirow{1}{*}{\textbf{Indictrans2}} 
                & \texttt{Direct}& 1.247& 0.74& 1.275& 0.77& 1.243& 0.769& 1.24& 0.747\\
            \midrule
            \multirow{2}{*}{\textbf{LLama-3.2-3B}}& \texttt{IdiomCE}& 1.34& 0.59& 1.2& 0.6& 1.105& 0.51& 1.18& 0.51\\
                & \texttt{Direct}& 1.12& 0.62& 1.05& 0.6& 1.04& 0.52& 1.07& 0.52\\
            \midrule
            \multirow{2}{*}{\textbf{Gemma2-9b-it}}& \texttt{IdiomCE}& \textbf{1.88}& 0.68& \textbf{1.7}& 0.68& \textbf{1.63}& 0.67& \textbf{1.56}& 0.62\\
                & \texttt{Direct}& 1.6& \textbf{0.73}& 1.44& \textbf{0.71}& 1.56& \textbf{0.71}& 1.46& \textbf{0.67}\\
            \midrule
            \multirow{2}{*}{\textbf{LLama-3.1-8B}}& \texttt{IdiomCE}& 1.655& 0.63& 1.40& 0.63& 1.25& 0.57& 1.3& 0.54\\
                & \texttt{Direct}& 1.27& 0.68& 1.23& 0.67& 1.16& 0.62& 1.12& 0.59\\
            \midrule
            \multirow{2}{*}{\textbf{GPT-4o}}& \texttt{IdiomCE}& 2.39& 0.70& 2.25& 0.69& 1.87& 0.67& 1.83& 0.66\\
                & \texttt{Direct}& 2.14& 0.73& 1.99& 0.764& 1.741& 0.72& 1.67&0.71\\
            \bottomrule
        \end{tabular}
    }
    \caption{Performance Metrics of Various Models on Mixed Dataset; COMET range [0,1].}
    \label{tab:mixed}
\end{table*}


\begin{table*}[ht]
    \centering
    \resizebox{\textwidth}{!}{
        \begin{tabular}{c  |c  |cc  |cc  |cc  |cc}
            \toprule
            \multirow{2}{*}{\textbf{Model}} & \multirow{2}{*}{\textbf{Methods}} & 
            \multicolumn{2}{c|}{{\texttt{hi-xx}}} & 
            \multicolumn{2}{c|}{{\texttt{bn-xx}}} & 
            \multicolumn{2}{c|}{{\texttt{ta-xx}}} & 
            \multicolumn{2}{c}{{\texttt{te-xx}}} \\
            \cmidrule(lr){3-4} \cmidrule(lr){5-6} \cmidrule(lr){7-8} \cmidrule(lr){9-10}
            & & \textbf{LLM-eval}& \textbf{COMET} & \textbf{LLM-eval}& \textbf{COMET} & \textbf{LLM-eval}& \textbf{COMET} & \textbf{LLM-eval}& \textbf{COMET} \\
            \midrule
            \multirow{1}{*}{\textbf{NLLB-200}} 
                & \texttt{Direct}& 1.85& 0.79& 1.70& 0.78& 1.84& 0.77& 1.81& 0.78\\
            \midrule
            \multirow{1}{*}{\textbf{Indictrans2}} 
                & \texttt{Direct}& \textbf{1.92}& 0.81& 1.78& 0.81& \textbf{2.01}& 0.77& 1.97& 0.77\\
            \midrule
            \multirow{2}{*}{\textbf{LLama-3.2-3B}}& \texttt{IdiomCE}& 1.263& 0.5663& 1.23& 0.5867& 1.2567& 0.53867& 1.273& 0.5493\\
                & \texttt{Direct}& 1.1867& 0.589& 1.17& 0.6163& 1.253& 0.572& 1.1867& 0.609\\
            \midrule
            \multirow{2}{*}{\textbf{Gemma2-9b-it}}& \texttt{IdiomCE}& \textbf{1.8233}& 0.7283& \textbf{1.783}& 0.727& \textbf{1.9867}& 0.7267& \textbf{2.02}& 0.724\\
                & \texttt{Direct}& 1.4833& \textbf{0.75}& 1.49& \textbf{0.775}& 1.563& \textbf{0.755}& 1.5467& \textbf{0.773}\\
            \midrule
            \multirow{2}{*}{\textbf{LLama-3.1-8B}}& \texttt{IdiomCE}& 1.42& 0.616& 1.46& 0.6404& 1.533& 0.5993& 1.493& 0.626\\
                & \texttt{Direct}& 1.34& 0.6533& 1.367& 0.688& 1.25& 0.6393& 1.25& 0.677\\
            \bottomrule
        \end{tabular}
    }
    \caption{Performance Metrics of Various Models For Inter-Indic languages; COMET range [0,1].}
    \label{tab:interindic}
\end{table*}

\section{Results}

\noindent {\bfseries Results on Mixed Dataset.} 
This dataset contains a mix of idioms, both seen and unseen during training. We conducted experiments on English-to-Hindi, Bengali, Tamil, and Telugu translation directions.
\noindent The results in Table \ref{tab:mixed} show: 1) IdiomCE, our approach that retrieves target idioms based on English idioms, outperforms the direct prompting method, highlighting the effectiveness of our retrieval-based training for idiomatic translation.
2) Among smaller models, Gemma2 9B achieves the best performance, even with direct prompting, demonstrating its strong capabilities in idiomatic translation.
3) With IdiomCE, very small models like Llama 3.2 3B perform comparably to the Directly Prompted larger Llama 3.1 8B variant.
4) Even for larger models like GPT-4o, IdiomCE improves performance, proving its effectiveness across different model sizes.
5) Foundational models like NLLB and IndicTrans2 struggle with idiomatic translation, showing low scores in LLM-eval.
On average, IdiomCE improves LLM-eval scores by 18.51\% for en-hi, 14.71\% for en-bn, 6.45\% for en-ta, and 10.33\% for en-te. We have also provided example translation in Appendix \ref{sec:examples}. Results on Additional baselines such as \cite{donthi-etal-2025-improving} are included in Appendix~\ref{tab:seen_res}.

\noindent {\bfseries Results on Seen and Unseen Dataset.} 
In Figure \ref{fig:unseen-seen-graph}, we have shown on average LLM evaluation for different models on various methods across languages. Notably, results for the IdiomKB baseline are shown only for the seen dataset, as IdiomKB supports only idioms present in the training set. On average, the Gemma2 9B model demonstrates the best performance among open-source LLMs on both seen and unseen datasets. Compared to IdiomKB and Direct Method, our approach, IdiomCE, outperform them by 14.28\% and 21.78\%, respectively, across open-source LLMs for seen dataset. Similarly, for unseen dataset, IdiomCE achieves 5.67\% improvement over direct method. Even with GPT-4o results, our approach shows significant improvements for both seen and unseen datasets. Further details on language-specific performance can be found in the Appendix in Table \ref{tab:seen_res} and \ref{tab:unseen_res}.

\noindent {\bfseries Results on Inter-Indic Languages.} 
Table \ref{tab:interindic} presents the average performance across Indic languages. Our findings indicate: 1)Using English as a pivot to retrieve idioms for translation between Indic languages improves LLM performance compared to direct prompting, highlighting the flexibility of our approach.
2) Gemma2 9B consistently performs well in inter-Indic translation settings, significantly outperforming other LLMs.
3) Interestingly, in some language pairs like hi-xx and ta-xx, IndicTrans2 achieves strong results, even surpassing other models.
Overall, IdiomCE demonstrates significant improvements in LLM evaluation, with a 12.5\% performance gain for hi-xx, 11.2\% for bn-xx, 17.5\% for ta-xx, and 19.9\% for te-xx translations over \texttt{Direct} prompting.

\begin{table}
    \centering
    \begin{tabular}{|c|c|c|c|}\hline
         Methods&  en-hi&  en-bn& en-tl\\\hline
         IdiomCE&  \textbf{3.51}&  \textbf{3.17}& \textbf{2.43}\\\hline
         IdiomKB&  2.65&  1.82& 1.88\\\hline
         Direct&  2.05&  1.58& 1.45\\ \hline
    \end{tabular}
    \caption{Human Evaluation on Idiomatic Translation on different methods.}
    \label{tab:human_eval}
\end{table}

\noindent{\bfseries{IdiomCE performance under Human evaluation.}} To compare the performance of IdiomCE with existing baselines, we conducted a manual quality annotation of translations generated by IdiomCE, IdiomKB, and direct translations from Gemma2-9b-it, as this model demonstrated superior performance across the evaluated methods (see Table \ref{tab:mixed}). The evaluation involved 19 native speakers who are highly fluent and bilingual. Assessments were carried out across three language pairs: English–Hindi (en-hi), English–Bengali (en-bn), and English–Telugu (en-tl). Each evaluator was presented with a source sentence containing idiomatic expressions and three corresponding translations produced by the different systems. Evaluators rated each translation on a 5-point scale, with detailed scoring criteria provided in the Appendix \ref{sec:Appendix}. As shown in Table \ref{sec:human_ins}, IdiomCE consistently outperformed the other baselines across all three language pairs. The performance gap was especially pronounced in the en-hi and en-bn directions, suggesting that the model is more effective at leveraging GNN-retrieved context for Hindi and Bengali than for Telugu. 

\noindent{\bfseries{Error Analysis.}} In addition to the human evaluation, we performed an error analysis to identify potential areas for improvement in our methodology. Upon examining the translations, we categorized the observed errors into three distinct types, as outlined below:
\begin{itemize}
    \item \textbf{Morphological Issues.} In some cases, Llama 3.1 8B and Llama 3.2 3B directly replaced an idiom without adapting its morphology, leading to unnatural phrasing in the target language. This suggests that smaller models struggle with idiom adaptation, whereas larger models perform better by adjusting idiomatic structures to fit grammatical norms. These observations highlight scalability challenges in idiomatic translation for smaller models, emphasizing the need for additional fine-tuning or external knowledge integration for improved performance.
    \item \textbf{Incorrect Selection.} In smaller models, such as LLaMA 3.2 3B, the model struggles to correctly select the appropriate target idiom for translation. This issue persists even when the GNN Top-K retrieval includes high-quality idiomatic translations. We have observed this phenomenon more frequently in languages such as Tamil and Telugu.
    \item \textbf{Pivot Noise.} For inter-Indic translations, we employ English as a pivot language to facilitate translation from one Indic language to another, leveraging the bidirectional property of GNN. However, this approach introduces potential noise, which can result in the best target-language idioms ranking lower in the Top-K retrieval. In some cases, high-quality idiomatic translations are entirely excluded from the retrieved set, leading to inaccuracies in the final translation.
\end{itemize}


\begin{table}
    \centering
    \resizebox{0.45\textwidth}{!}{
    \begin{tabular}{|c|c|c|}
        \hline
        Hits @k & Without NodeDup & With NodeDup \\
        \hline
        Hits@5 & 81.33 $\pm$ 2.36 & \textbf{85.28} $\pm$ 2.99 \\
        \hline
        Hits@10 & 90.00 $\pm$ 2.36 & \textbf{96.28}$\pm$ 1.37 \\
        \hline
        Hits@20 & 100.00 $\pm$ 0.00 & 100.00 $\pm$ 0.00 \\
        \hline
        Hits@50 & 100.00 $\pm$ 0.00 & 100.00 $\pm$ 0.00 \\
        \hline
        AUC & 95.32 & \textbf{96.33} \\
        \hline
    \end{tabular}
    }
    \caption{Ablation on Node Duplication module.}
    \label{tab:ablation-nodedup}
\end{table}

\noindent{\bfseries{Ablation Studies.}}
To justify the use of the Node Duplication procedure (see Sec 3.2), we conduct an ablation experiment comparing performance with and without the NodeDup module in Table \ref{tab:ablation-nodedup}. We report Hits@k \cite{hits} for the en-hi translation task, which includes 8,233 nodes (~4.6K Hindi target nodes), with 1.1K cold target nodes. Our results show that incorporating the NodeDup module improves Hits@k by 4.85\% for $k=5$ and 6.97\% for $k=10$, demonstrating its effectiveness in enhancing target node retrieval.



\section{Conclusion}
In this work, we addressed the challenges of idiomatic translation by introducing IdiomCE, an adaptive GNN-based approach that effectively captures the complex relationships between idiomatic expressions across languages. Our method generalizes to seen and unseen idioms, improves translation quality even in smaller models, and enables translation via a pivot language through the GNN framework. Experimental results across multiple Indian languages demonstrate that our approach outperforms traditional static knowledge graphs and prompt-based methods, significantly improving idiomatic translation. By leveraging GPT-4 as an evaluation metric, we show that our model better preserves meaning and cultural nuances in translation. Future work can extend this approach to more languages and richer contextual signals.

\section*{Limitations}
While our work shows significant improvements in idiomatic translation, we mention some of the limitations of our work. Our approach heavily depends on the synthetically generated cultural elements (features). Noisy features, especially in low-resource languages, might affect the performance of our method. Secondly, as mentioned before, although our model captures idiomatic mappings, some idioms rely heavily on a deep contextual understanding of the surrounding sentences and not just on the training data used, which can limit the model's performance.



\bibliography{acl-arxiv}
\newpage
\appendix
\label{sec:Appendix}
\section{Prompts used in the experiments}
\label{sec:Prompt}
\subsection{Direct Prompt}
\label{sec:direcet_prompt}
\begin{figure}[h]
    \vspace{-3mm} 
    \centering
    \includegraphics[width=\linewidth]{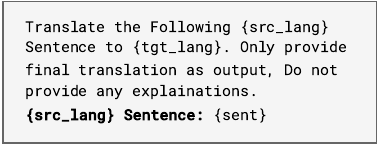}
    \vspace{-10mm}
    \label{fig:direct_prompt}
\end{figure}

\subsection{Selection Prompt}
\label{sec:Selection_prompt}
\begin{figure}[h]
    \vspace{-3mm} 
    \centering
    \includegraphics[width=\linewidth]{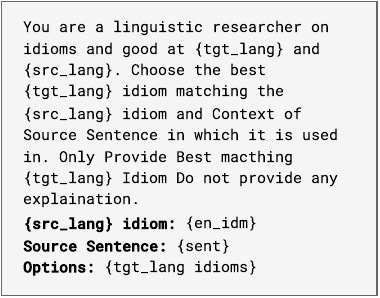}
    \vspace{-10mm}
    \label{fig:selection_prompt}
\end{figure}

\subsection{Translation Prompt}
\label{sec:Translation_prompt}
\begin{figure}[!htb]
    \vspace{-3mm} 
    \centering
    \includegraphics[width=\linewidth]{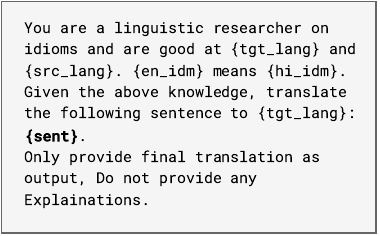}
    \vspace{-10mm}
    \label{fig:translation_prompt}
\end{figure}

\newpage
\subsection{Cultural element generation prompt}
\label{sec:Generation_prompt}
\begin{figure}[h]
    \vspace{-3mm} 
    \centering
    \includegraphics[width=\linewidth]{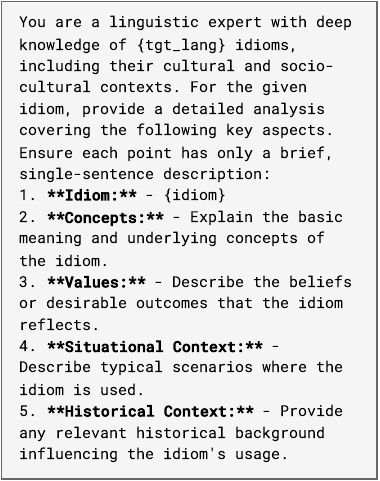}
    \vspace{-5mm}
    \label{fig:generation_prompt}
\end{figure}

\noindent {\bfseries Training Details.} 
We train the GNN using a 2-layer SAGEConv architecture, mapping input states from 768 to a hidden dimension of 64. The hidden representation then passes through an MLP with two linear layers and ReLU activation. The model is trained for 50 epochs over 5 runs.
For Node Duplication Augmentation, each target node is duplicated twice, and the distinction threshold ($\delta$) between cold and warm nodes is set to 3. $\alpha$ used as margin for Contrastive Training is set to 1. \\

\label{sec:human_ins}
\noindent {\bfseries Human Evaluation Instruction.} Here we provide details on Human Evaluation Instruction,  we conducted human evaluation on three translation directions: English to Hindi, English to Bengali, and English to Telugu. The evaluation compared three methods: IdiomCE (ours), IdiomKB, and Direct Prompting results are shared in \ref{tab:human_eval}. A total of 19 native speakers, who are highly fluent and bilingual, participated in the evaluation. Each evaluator was presented with: 1) A source sentence containing an idiom. 2) Three translations generated by the different methods. Instructions to score the translations on a scale of 1 to 5, based on the following criteria:
\begin{itemize}
    \item \textbf{Score 1:} The sentence is correctly translated, but the idiom is completely mistranslated, missing its figurative meaning or translated literally.
    \item \textbf{Score 2:} The sentence is correctly translated, and the idiom is translated, but it does not fully convey the intended meaning.
    \item \textbf{Score 3:} The sentence and idiom are correctly translated, but the idiomatic expression does not sound natural to native speakers.
    \item \textbf{Score 4:} The sentence and idiom are accurately translated, highly natural, and the overall translation is fluent.
    \item \textbf{Score 5:} The translation is perfectly natural for native speakers, with the idiom translated in the best possible way. This evaluation provides insights into how well each method captures idiomatic expressions while maintaining fluency and naturalness.
\end{itemize}

\label{sec:data_comp}
\noindent {\bfseries Dataset Composition.} The total number of unique idioms per language in our training dataset is as follows: Telugu - 4,407, Bengali - 4,479, Tamil - 4,179, Hindi - 4,722, and English – 4,500. When this data is transformed into a graphical structure, the training dataset—prior to applying the Node Duplication Augmentation strategy the Training Composition for GNN expands to:
\begin{itemize}
    \item English and Tamil: 7,646
    \item English and Telugu: 7,988
    \item English and Bengali: 7,872
    \item English and Hindi: 8,233
\end{itemize}

The test set, as detailed in Section 4 (Datasets), it consists of 400 unique idiom-containing sentences, with 200 sentences featuring seen idioms and 200 sentences containing unseen idioms. This is referred to as the Mixed Dataset in our paper. Additionally, for the inter-Indic translation setting, we include 200 sentences, each containing a unique idiom. \\

\label{sec:time_comp}
\noindent {\bfseries Computational Resource and Inference time.} We thank the reviewer for their suggestion. Below is a detailed breakdown of the computational resources used: GPU Specifications: The experiments were conducted on an NVIDIA GeForce RTX 3090 with CUDA version 12.2 and 24GB of VRAM. Training Time: Training the GNN for a single language direction takes approximately 6–7 minutes, with a total GPU memory requirement of around 500MB. Inference Time and GPU utilization:
\begin{itemize}
    \item On the seen dataset (200 sentences): 14–15 minutes.
    \item On the unseen dataset (200 sentences): 20–22 minutes.
    \item For the inter-Indic translation setting (200 sentences): 17–18 minutes.
\end{itemize}
Overall GPU Utilization: When the entire inference pipeline runs simultaneously, the maximum GPU memory required is approximately 10–11GB.

\newpage
\section{Examples}
\label{sec:examples}
\subsection{Translation Example \texttt{en-hi} direction}
\begin{figure}[ht]
    \vspace{-2mm} 
    \centering
    \includegraphics[width=1\linewidth, height=9.5cm, keepaspectratio]{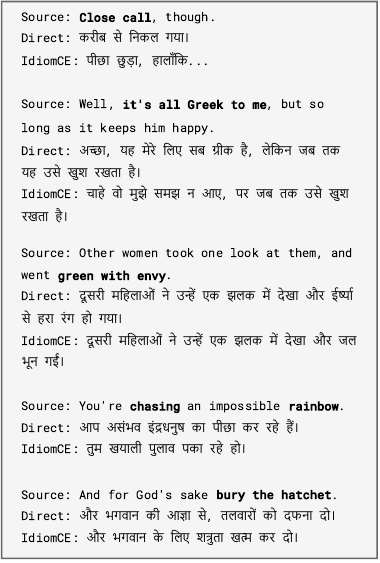}
    \label{fig:hindigen}
\end{figure}

\subsection{Translation Example \texttt{en-bn} direction}
\begin{figure}[h]
    \vspace{-3mm} 
    \centering
    \includegraphics[width=\linewidth, height=10.5cm, keepaspectratio]{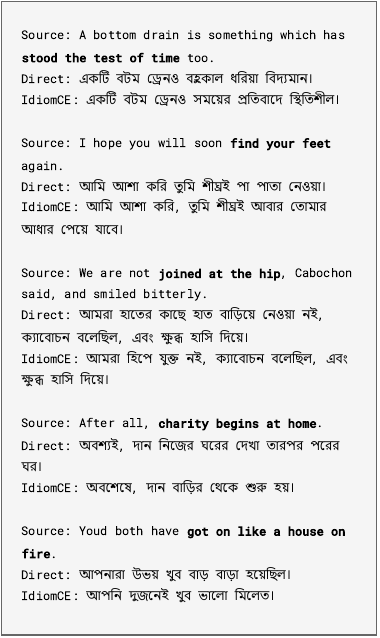}
    \vspace{-10mm}
    \label{fig:bengaligen}
\end{figure}

\newpage
\subsection{Translation Example \texttt{en-ta} direction}
\begin{figure}[h]
    \vspace{-3mm} 
    \centering
    \includegraphics[width=\linewidth]{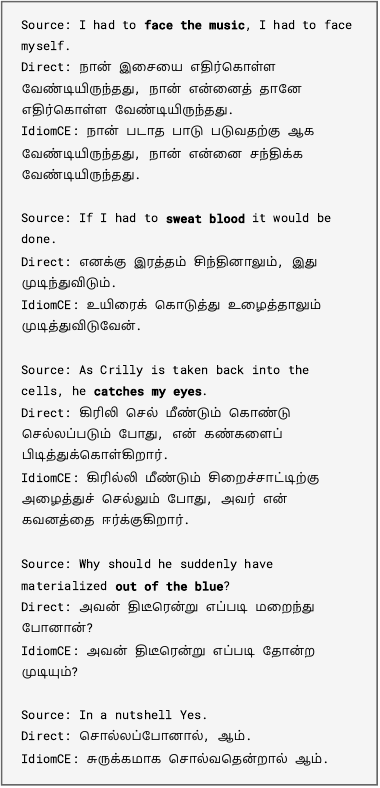}
    \vspace{-10mm}
    \label{fig:tamilgen}
\end{figure}

\newpage
\subsection{Translation Example \texttt{en-te} direction}
\begin{figure}[h]
    \vspace{-3mm} 
    \centering
    \includegraphics[width=\linewidth]{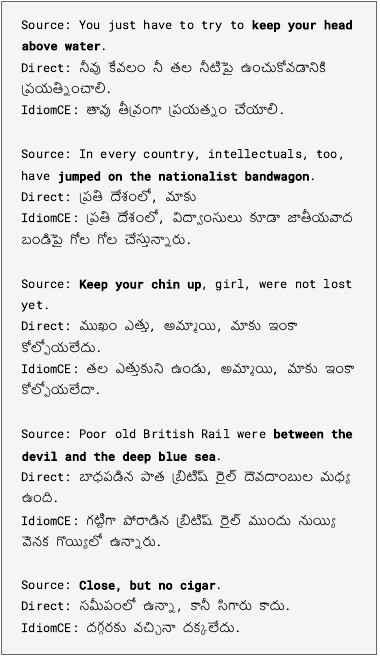}
    \vspace{-10mm}
    \label{fig:telugugen}
\end{figure}

\begin{table*}[ht]
    \centering
    \resizebox{\textwidth}{!}{
        \begin{tabular}{c|c|c|c|c|c}
            \toprule
            {Language} & \texttt{Hits@5} & \texttt{Hits@10} & \texttt{Hits@20}  & \texttt{Hits@50}  & \texttt{AUC} \\
            \midrule
            \texttt{hindi}      & 85.28 $\pm$ 2.99& 96.28 $\pm$ 1.37& 100.00 $\pm$ 0.00  & 100.00 $\pm$ 0.00  & 96.33 $\pm$ 0.28\\
            \texttt{Telugu}  & 82.50 $\pm$ 8.54  & 95.83 $\pm$ 2.95  & 100.00 $\pm$ 0.00 & 100.00 $\pm$ 0.00  & 95.32 $\pm$ 0.37 \\
            \texttt{Tamil}   & 76.06 $\pm$ 3.98  & 88.45 $\pm$ 2.09  & 98.59 $\pm$ 1.00  & 100.00 $\pm$ 0.00  & 93.27 $\pm$ 0.73 \\
            \texttt{Bengali} & 79.29 $\pm$ 5.30  & 95.00 $\pm$ 4.07  & 99.29 $\pm$ 1.60  & 100.00 $\pm$ 0.00  & 96.10 $\pm$ 0.12 \\
            \bottomrule
        \end{tabular}
    }
    \caption{Performance on GNN Link Prediction task for each language.}
    \label{tab:gnn_performance}
\end{table*}

\begin{table*}[ht]
    \centering
    \resizebox{\textwidth}{!}{
        \begin{tabular}{c  |c  |cc  |cc  |cc  |cc}
            \toprule
            \multirow{2}{*}{\textbf{Model}} & \multirow{2}{*}{\textbf{Methods}} & 
            \multicolumn{2}{c|}{{\texttt{en-hi}}} & 
            \multicolumn{2}{c|}{{\texttt{en-bn}}} & 
            \multicolumn{2}{c|}{{\texttt{en-ta}}} & 
            \multicolumn{2}{c}{{\texttt{en-te}}} \\
            \cmidrule(lr){3-4} \cmidrule(lr){5-6} \cmidrule(lr){7-8} \cmidrule(lr){9-10}
            & & \textbf{GPT-4} & \textbf{COMET} & \textbf{GPT-4} & \textbf{COMET} & \textbf{GPT-4} & \textbf{COMET} & \textbf{GPT-4} & \textbf{COMET} \\
            \midrule
            \multirow{1}{*}{\textbf{NLLB-200}} 
                & \texttt{Direct}& 1.34& 0.70& 1.45& 0.77& 1.21& 0.69& 1.14& 0.64\\
            \midrule
            \multirow{1}{*}{\textbf{Indictrans2}} 
                & \texttt{Direct}& 1.24& 0.74& 1.27& 0.78& 1.26& 0.76& 1.21& 0.74\\
            \midrule
            \multirow{5}{*}{\textbf{LLama-3.2-3B}}& \texttt{IdiomCE}& 1.42& 0.58& 1.26& 0.59& 1.15& 0.52& 1.24& 0.51\\
                & \texttt{Direct}& 1.12& 0.62& 1.06& 0.60& 1.03& 0.51& 1.09& 0.54\\
 & IdiomKB& 1.25& 0.61& 1.05& 0.59& 1.07& 0.52& 1.11&0.52\\
 & LIA& 1.13& 0.565& 1.01& 0.5702& 0.97& 0.510& 1.06&0.491\\
 & SIA& 1.18& 0.57& 1.15& 0.58& 1.10& 0.53& 1.09&0.48\\
            \midrule
            \multirow{5}{*}{\textbf{Gemma2-9b-it}}& \texttt{IdiomCE}& 2.08& 0.69& 1.84& 0.69& 1.76& 0.68& 1.68& 0.63\\
                & \texttt{Direct}& 1.63& 0.73& 1.50& 0.71& 1.60& 0.72& 1.45& 0.68\\
 & IdiomKB& 1.875& 0.70& 1.64& 0.70& 1.65& 0.68& 1.50&0.64\\
 & LIA& 1.30& 0.64& 1.18& 0.610& 1.184& 0.554& 1.125&0.517\\
 & SIA& 1.41& 0.65& 1.30& 0.63& 1.23& 0.562& 1.20&0.55\\
            \midrule
            \multirow{5}{*}{\textbf{LLama-3.1-8B}}& \texttt{IdiomCE}& 1.89& 0.62& 1.54& 0.63& 1.29& 0.57& 1.41& 0.53\\
                & \texttt{Direct}& 1.27& 0.68& 1.22& 0.67& 1.16& 0.62& 1.14& 0.58\\
 & IdiomKB& 1.40& 0.67& 1.19& 0.67& 1.20& 0.60& 1.21&0.59\\
 & LIA& 1.60& 0.645& 1.50& 0.632& 1.42& 0.61& 1.35&0.56\\
 & SIA& 2.08& 0.69& 1.84& 0.69& 1.60& 0.65& 1.68&0.63\\
        \end{tabular}
    }
    \caption{Performance Metrics of Various Models on Seen Dataset; COMET range [0,1].}
    \label{tab:seen_res}
\end{table*}

\begin{table*}[ht]
    \centering
    \resizebox{\textwidth}{!}{
        \begin{tabular}{c  |c  |cc  |cc  |cc  |cc}
            \toprule
            \multirow{2}{*}{\textbf{Model}} & \multirow{2}{*}{\textbf{Methods}} & 
            \multicolumn{2}{c|}{{\texttt{en-hi}}} & 
            \multicolumn{2}{c|}{{\texttt{en-bn}}} & 
            \multicolumn{2}{c|}{{\texttt{en-ta}}} & 
            \multicolumn{2}{c}{{\texttt{en-te}}} \\
            \cmidrule(lr){3-4} \cmidrule(lr){5-6} \cmidrule(lr){7-8} \cmidrule(lr){9-10}
            & & \textbf{LLM-eval}& \textbf{COMET} & \textbf{LLM-eval}& \textbf{COMET} & \textbf{LLM-eval}& \textbf{COMET} & \textbf{LLM-eval}& \textbf{COMET} \\
            \midrule
            \multirow{1}{*}{\textbf{NLLB-200}} 
                & \texttt{Direct}& 1.26& 0.70& 1.41& 0.77& 1.17& 0.69& 1.06& 0.64\\
            \midrule
            \multirow{1}{*}{\textbf{Indictrans2}} 
                & \texttt{Direct}& 1.25& 0.74& 1.28& 0.78& 1.22& 0.76& 1.27& 0.74\\
            \midrule
            \multirow{3}{*}{\textbf{LLama-3.2-3B}}& \texttt{IdiomCE}& 1.25& 0.58& 1.14& 0.59& 1.06& 0.52& 1.13& 0.51\\
 & LIA& 1.13& 0.564& 1.05& 0.570& 1.06& 0.508& 1.02&0.491\\
                & \texttt{Direct}& 1.12& 0.62& 1.05& 0.60& 1.05& 0.51& 1.05& 0.53\\
            \midrule
            \multirow{3}{*}{\textbf{Gemma2-9b-it}}& \texttt{IdiomCE}& 1.68& 0.68& 1.56& 0.67& 1.50& 0.68& 1.49& 0.66\\
 & LIA& 1.34& 0.62& 1.17& 0.610& 1.11& 0.56& 1.12&0.517\\
                & \texttt{Direct}& 1.57& 0.72& 1.39& 0.70& 1.53& 0.72& 1.4& 0.68\\
            \midrule
            \multirow{3}{*}{\textbf{LLama-3.1-8B}}& \texttt{IdiomCE}& 1.42& 0.63& 1.27& 0.62& 1.21& 0.59& 1.19& 0.53\\
 & LIA& 1.61& 0.65& 1.50& 0.653& 1.40& 0.64& 1.38&0.587\\
                & \texttt{Direct}& 1.28& 0.68& 1.23& 0.67& 1.16& 0.62& 1.11& 0.55\\
        \end{tabular}
    }
    \caption{Performance Metrics of Various Models on Unseen Dataset; COMET range [0,1].}
    \label{tab:unseen_res}
\end{table*}

\end{document}